\documentclass{article}

\usepackage[preprint, nonatbib]{neurips_2024}

\usepackage[utf8]{inputenc} % allow utf-8 input
\usepackage[T1]{fontenc}    % use 8-bit T1 fonts
\usepackage{url}            % simple URL typesetting
\usepackage{booktabs}       % professional-quality tables
\usepackage{amsfonts}       % blackboard math symbols
\usepackage{nicefrac}       % compact symbols for 1/2, etc.
\usepackage{microtype}      % microtypography
\usepackage{xcolor}         % colors
\usepackage{graphicx}
\usepackage{amsmath,amssymb,newtxmath}
\usepackage{makecell}
\usepackage{caption}
\usepackage{wrapfig}
\usepackage{adjustbox}
\usepackage{colortbl}

\usepackage{color, colortbl}
\definecolor{citecolor}{HTML}{2980b9}
\definecolor{linkcolor}{HTML}{c0392b}

\usepackage{hyperref}
\hypersetup{hidelinks,breaklinks=true,colorlinks,citecolor=citecolor,linkcolor=linkcolor}

\newcommand{\squishlist}{
 \begin{list}{$\bullet$}
  { \setlength{\itemsep}{0pt}
     \setlength{\parsep}{1pt}
     \setlength{\topsep}{1pt}
     \setlength{\partopsep}{0pt}
     \setlength{\leftmargin}{1.5em}
     \setlength{\labelwidth}{1em}
     \setlength{\labelsep}{0.5em} } }
\newcommand{\squishend}{
  \end{list}  }

\title{NoiseBoost: Alleviating Hallucination with Noise Perturbation for Multimodal Large Language Models}

\author{%
    Kai WU$^1$\thanks{Equal contribution.} \hspace{0.3cm} 
    Boyuan Jiang$^1$\footnotemark[\value{footnote}] \hspace{0.3cm} 
    Zhengkai Jiang$^1$\\
    \textbf{Qingdong He}$^1$ \hspace{0.3cm} 
    \textbf{Donghao Luo}$^1$ \hspace{0.3cm} 
    \textbf{Shengzhi Wang}$^2$ \hspace{0.3cm} 
    \textbf{Chengjie Wang}$^1$ \hspace{0.3cm} 
    \textbf{Qingwen Liu}$^2$\thanks{Corresponding author.} \\
  Tencent Youtu Lab \hspace{0.3cm} Tongji University
}

\begin{document}

\maketitle

\begin{abstract}
% Large Vision Language Model (MLLM) provides an effective way to understand visual information through large language models.
% To facilitate direct comprehension, recent works apply no image augmentations during visual processing, resulting in less training diversity of images than language. 
% This imbalance leads to MLLMs' overdependence on linguistic data, resulting in hallucinations and a neglect of the vision information. 
% In this paper, we propose NoiseBoost, a well-generalized method that enhances MLLMs by integrating noise perturbations into the visual features. 
% The feature noise injection serves as a data augmentation technique without incurring additional training costs.
% Despite its simplicity, NoiseBoost consistently elevates MLLM performance across a variety of tasks—including Image Captioning, hallucination mitigation, and Visual Question Answering—as well as prevalent training methodologies such as supervised fine-tuning, reinforcement learning, and the teacher-student model.
% This study provides valuable insights into the role of noise perturbations in enhancing the robustness and performance of MLLM, paving the way for more effective and efficient multimodal learning systems.
Multimodal large language models (MLLMs) contribute a powerful mechanism to understanding visual information building on large language models.
However, MLLMs are notorious for suffering from hallucinations, especially when generating lengthy, detailed descriptions for images. 
Our analysis reveals that hallucinations stem from the inherent summarization mechanism of large language models, leading to excessive dependence on linguistic tokens while neglecting vision information. 
In this paper, we propose NoiseBoost, a broadly applicable and simple method for alleviating hallucinations for MLLMs through the integration of noise feature perturbations.
Noise perturbation acts as a regularizer, facilitating a balanced distribution of attention weights among visual and linguistic tokens. 
Despite its simplicity, NoiseBoost consistently enhances the performance of MLLMs across common training strategies, including supervised fine-tuning and reinforcement learning. 
Further, NoiseBoost pioneerly enables semi-supervised learning for MLLMs, unleashing the power of unlabeled data.
Comprehensive experiments demonstrate that NoiseBoost improves dense caption accuracy by 8.1\% with human evaluation and achieves comparable results with 50\% of the data by mining unlabeled data. Code and models are available at https://kaiwu5.github.io/noiseboost.
% This study provides valuable insights into the role of noise perturbations in enhancing the robustness and performance of MLLMs, paving the way for more effective and efficient multimodal learning systems.
\end{abstract}

\section{Introduction}

% Image describing ability usually resembles how humans understand the world. 
% With the recent development of Large Language vision models, we delicately teach that ability to AI which can densely describe an image with hundreds of words. 
% However, 
% Large language models (LLMs), such as GPT-4, have demonstrated significant potential in approximating human intelligence, attributable to the scaling laws by expanding training datasets and computational parameters. 
Recent Large language models (LLMs)~\cite{achiam2023gpt,touvron2023llama,touvron2023llama2,team2024gemma} have demonstrated significant potential in approximating human intelligence and can serve as sophisticated assistants for intricate tasks. 
% , including solving mathematical problems, organizing travel plans, or responding to inquiries. 
% Genuinely general-purpose assistants must be capable of interfacing with the natural world by processing both linguistic and visual information~\cite{hong2023cogagent}, thereby mirroring human behaviour in the execution of real-world tasks.
Building on the foundational LLMs, Multimodal Large Language Models (MLLMs)~\cite{liu2024LLaVA, liu2023LLaVA1.5,bai2023qwenvl} are designed to transfer LLM's zero-shot understanding ability to vision, extending the advantages of LLMs to the realm of multi-modality comprehension.
Despite the significant progress made in recent MLLM research, no MLLM method can be immune to hallucinations~\cite{yu2023hallucidoctor,liu2023mitigatinghallu} which limits their applicability in real-world applications.

\begin{figure}
  \centering
  \includegraphics[width=1.0\linewidth]{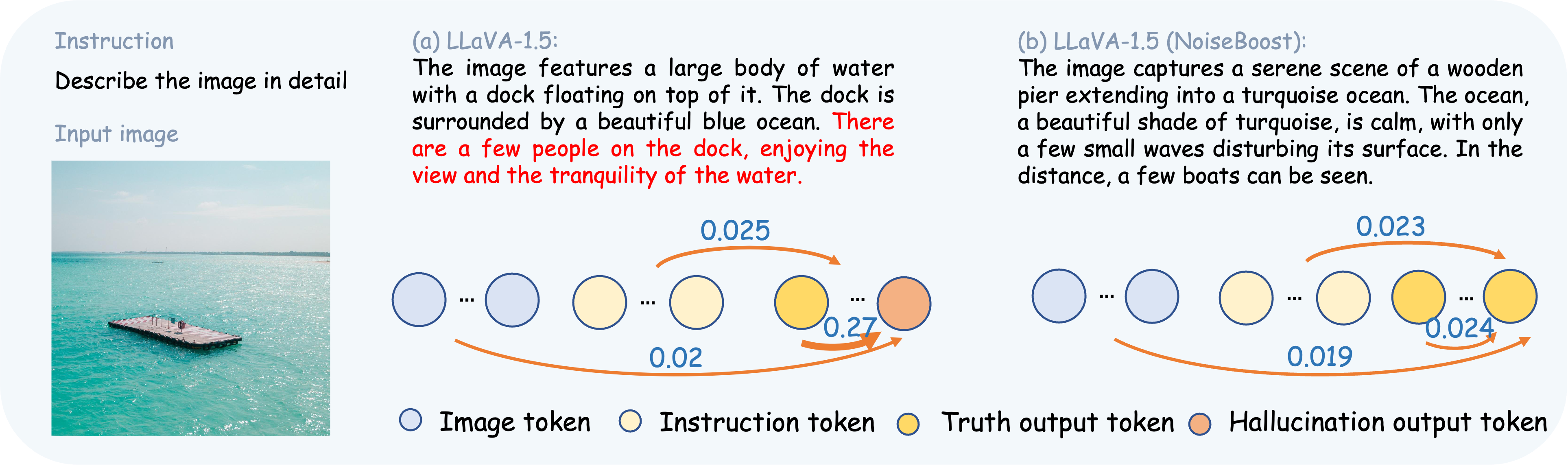}
  \caption{MLLMs suffer from hallucinations due to the over-reliance on language priors. In (a), the hallucination tokens are overly dependent (0.27) on previous language tokens, and later tokens are all hallucinations. 
Meanwhile, in (b), NoiseBoost helps MLLMs distribute the attention weights evenly among visual and language tokens by noise perturbation, leading to honest results.}
  \label{fig:intuition}
\end{figure}

Recent studies on mitigating hallucination predominantly concentrate on the development of a tailored decoder or the annotation of hallucination-specific data. 
OPERA~\cite{huang2023opera}, utilizing a discovery and re-decoding loop, implements a language over-trust penalty to discard hallucinated results and progressively regenerate them.
By introducing visual contrastive decoding, ~\cite{leng2023mitigatingContrast} subtracts the decoded logits of hallucinated visual inputs from those of the original input. 
% This approach operates under the assumption that distorted visual inputs are more likely to induce hallucinations. 
% The subtraction operation aims to recalibrate an excessive reliance on language priors. 
Despite these re-decoding-based methods do not require training, they achieve performance improvements by doubling or even tripling the inference time, rendering MLLMs challenging for deployment on personal devices.
Conversely, HallDoctor~\cite{yu2023hallucidoctor,pi2024strengtheningHallBoost} establish a preference dataset annotation pipeline that distorts the ground truth answer with errors to form hallucinated pairs, which is later trained to align the model's honesty. 
Without distorting the training response, Fine-grained PPO~\cite{wu2024fineGPPO} annotates the model response on a word-by-word basis to train a reward model and align the model's generation with proximal policy optimization~\cite{ouyang2022instructgpt}. However, these manually curated reward datasets differ in distribution from real-world usage and cannot encompass all scenarios.
% , given that LLaMA3 trains on more than 15 trillion tokens from the web corpus. 
In this paper, we aim to identify the fundamental reasons for hallucination and enhance MLLM's training without additional datasets or training costs.

Upon diving into the attention mechanism of MLLMs, we discovered that the occurrence of hallucination could be attributed to an excessive dependence on language priors.
During the LLM response generation process, certain language tokens are automatically selected as anchors~\cite{pang2024anchor}, causing subsequent generations to rely more heavily on the summarization of anchor token information, rather than on the comprehensive set of preceding visual and linguistic tokens in the context. 
% This pattern triggers the in-context learning of the LLM~\cite{alayrac2022flamingo}, facilitating the transfer of information from the example label words to the predicted responses. 
As depicted in Fig.~\ref{fig:intuition}(a), MLLM's information flow is unevenly distributed from visual and language tokens.
Furthermore, MLLM's visual and language tokens are from separately pre-trained visual encoder and LLMs~\cite{liu2024LLaVA,liu2023LLaVA1.5,bai2023qwenvl}, leading to a significant disparity in features even after training.
Since MLLM's anchor token selection frequency is correlated with generation length, the hallucination phenomenon gets worse when generating long, detailed descriptions.
% By visualizing the token correlation, the column highlighted in red demonstrates the reliance on summary tokens during MLLM decoding.
% Moreover, the summarizing behavior of the LLM is unpredictable and uncontrollable.
Without appropriate training methodologies, the flow of information from visual tokens to linguistic tokens is hindered, leading to a neglect of visual information and an over-reliance on language priors.
In this paper, we propose NoiseBoost, a simple and widely applicable noise perturbation method designed to mitigate hallucination across various MLLM training stages.
NoiseBoost disrupts the excessive dependence on language priors, facilitating a balanced distribution of the model's attention between visual and linguistic tokens.  
Specifically, we increase the hardship in MLLM's learning process by incorporating noise feature perturbation, achieved by injecting noise into visual tokens. 
This approach complicates visual understanding, necessitating more evenly distributed attention weights in LLM. 
Our extensive experiments demonstrate that the injection of Gaussian noise to projected visual tokens consistently enhances performance with negligible additional training costs.
As depicted in Fig.\ref{fig:intuition}(b), token correlation is evenly distributed, significantly reducing overconfidence induced by summary tokens in LLMs. 
To further exhibit NoistBoost's generalizability, we conducted experiments across two MLLM training stages: supervised fine-tuning and reinforcement learning. 
NoiseBoost consistently improves performance in both training methods across hallucination and question-answer datasets, validating the efficacy of feature perturbation. 
To verify the results of long description generations, we evaluated 1k images by annotators for dense captions, which NoiseBoost shows an 8.1\% improvement in accuracy.

By integrating NoiseBoost, we pioneer the incorporation of semi-supervised learning (SSL) architecture for MLLM models. 
Current MLLM training relies on noisy web corpus, incurring substantial labeling costs for data cleaning without harnessing the potential of unlabeled data. 
The challenge is that MLLM does not have a mechanism for teacher-student learning with pseudo labels, which is a crucial element in traditional SSL architecture. 
We generate pseudo labels using original images and use NoiseBoost to be the noisy student, learning to produce consistent and robust results. 
Our experiments show that NoiseBoost can unleash the power of unlabeled data and achieve similar performance with 50\% of labeled data. 

In summary, our contributions are as follows:

% \begin{itemize}
\squishlist
    \item With analyzing the cause of hallucination, We propose a simple and well-generalized method, NoiseBoost, which effectively alleviates hallucination for MLLM at negligible additional cost without introducing extra data. 
    \item We are the pioneers in facilitating semi-supervised learning for MLLMs with NoiseBoost and reach the same performance with 50\% of training data by mining the power of unlabeled data.
    \item Extensive experiments indicate the effectiveness of NoistBoost as a general training enhancement method, providing consistent performance improvement for MLLMs.
% \end{itemize}
\squishend

% Despite its simplicity, NoiseBoost effectively alleviates hallucination at negligible additional cost without introducing extra data. By incorporating NoiseBoost, MLLM can be trained with semi-supervised learning technologies and shed light on unlabeled data mining. With extensive experiments, NoiseBoost consistently improves on MLLMs with 2\% on supervised learning and 3\% on reinforcement learning and achieves similar performance with half the data. 
% \begin{itemize}
% \item We propose NoiseBoost, a novel feature perturbation method that can improve the performance of MLLMs by increasing vision diversity and generalizing well on diverse training strategies, including SFT, reinforcement learning, and semi-supervised learning. 
% \item Based on NoiseBoost, we conduct thorough quantitive and qualitative analyses to explain why standard augmentation methods are not effective and why feature perturbation is effective.
% % \item We release our codes and checkpoints associated with this paper, which can facilitate the reasoning analysis of the details of MLLMs. 
% % factual error, expand knowledge, and collect data is hard, multi-supervision is helpful
% % hallucination, anchor tokens, image augmentation/ noise perturbation
% % dpo training for more human 

% % 1. recognition tasks has augmentation which makes it more robust, noise feature perturbation more focus on image

% % 2. dense supervision for model knowledge scope expansion

% % 3. DPO training more prone to human 
% \end{itemize}

\section{Related Work}

\subsection{Multimodal Large Language Foundation Models}
Recent advancements in MLLMs research are primarily attributed to the evolution of large language models (LLMs). To integrate vision models with LLMs, existing MLLMs typically utilize lightweight layers such as QFormer~\cite{li2023blip} or linear projection~\cite{liu2024LLaVA}. Notably, LLaVA~\cite{liu2024LLaVA} integrates a vision encoder and an LLM to facilitate general-purpose visual and language understanding. This is achieved using multi-modal language-image instruction-following data, with the vision encoder designed to project image features into language token representations. MiniGPT-4~\cite{zhu2023minigpt} incorporates a pretrained ViT and Q-Former and an LLM for multi-modal generation and understanding. 
% InternVL~\cite{chen2023internvl} introduces a novel image-text alignment strategy during training. 
Mini-gemini~\cite{li2024mini} enhances multi-modal reasoning capabilities through high-resolution visual tokens, employing an additional visual encoder for high-resolution refinement.
However, directly bridging visual and language modalities causes hallucinations from over-reliance on the language priors. We propose NoiseBoost to redistribute attention weight to both visual and linguistic tokens by injecting feature perturbations to visual features.  
% However, MLLMs usually have hallucinations, especially for long, detailed description generation. We take one step deeper to analyze the cause of the hallucination, which stems from the language's prior over-reliance. 

\subsection{Hallucinations in MLLMs}
Hallucination in MLLMs has significantly impeded their usage in the real world, especially for tasks that rely on precise captions.
Previous works focus on two perspectives:  dataset construction and decoding schemes to alleviate the hallucination in MLLMs. For dataset construction, HallDocter~\cite{yu2023hallucidoctor} proposes a pipeline to annotate the hallucination dataset with the help of GPT4V. To enable reinforcement learning, ~\cite{zhao2023hadpo,zhao2023beyond} propose object substitution using GPT4V and labor checking to create a hallucinated response pair. 
However, rectifying large models like MLLM with small curated data is contrary to the scaling law. 
With decoder scheme optimization, ~\cite{jiang2023hallcontrast} proposes to achieve an un-hallucinated response by subtracting the hallucinated response decoded simultaneously using only language prompts.
OPERA~\cite{huang2023opera} proposes a penalty-based found and re-decoding method to reduce hallucinations. 
Although effective, decoding-based methods require iterative decoding, which incurs computational burden and impedes MLLM's deployment on personal devices. 
In this paper, we design a simple and well-generalized noise perturbation method for alleviating hallucinations without introducing additional datasets or inference costs. 

\section{Method}
In this section, we first introduce the preliminaries of MLLM in Sec.\ref{sec:preliminaries}. Then we show how NoiseBoost is applied to different MLLM training methods, including Supervised Fine-tuning in Sec.\ref{sec:supervised} and reinforcement Learning in Sec.~\ref{sec:reinforce}. Finally, we incorporate Semi-Supervised Learning into MLLM by using NoiseBoost in Sec.\ref{sec:semisup}.

\subsection{Preliminaries}
\label{sec:preliminaries}
\begin{figure}
    \centering
    \includegraphics[width=\linewidth]{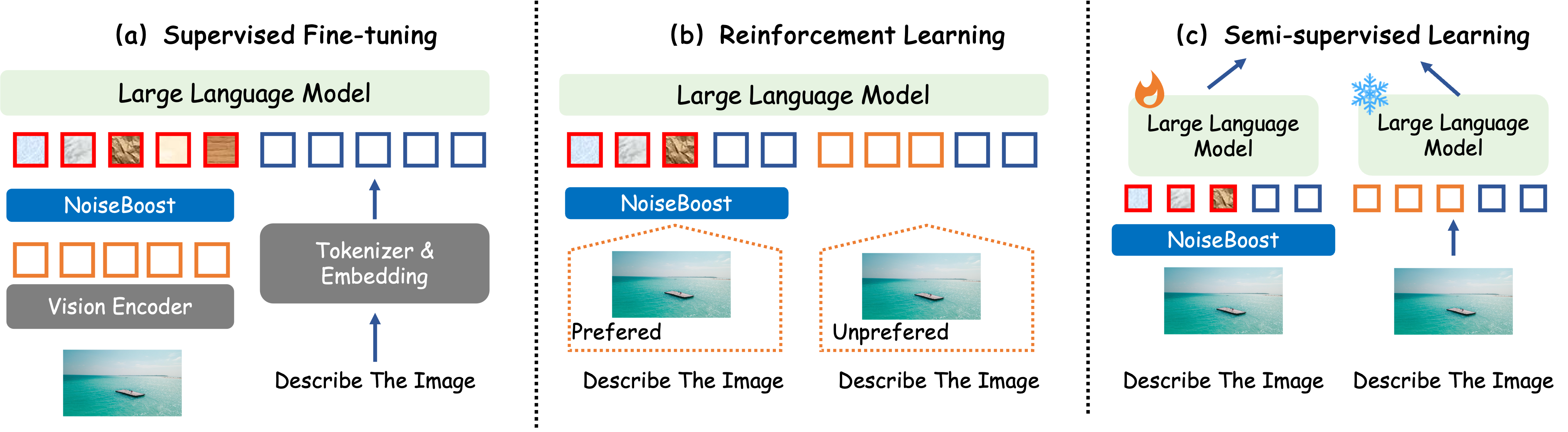}
    \caption{Framework of NoiseBoost. We add noise perturbation to visual tokens to mitigate the over-reliance on language tokens and thus fewer hallucinations. For SFT, we directly inject noise to visual features. We only inject perturbation to preferred response since that can make MLLMs harder to learn and achieve better results. For semi-supervised learning, we use freezed MLLM as a teacher to generate pseudo labels and NoiseBoost as students for consistency regularization. }
    \label{fig:framework}
\end{figure}

Multimodal large language models (MLLMs) attain visual comprehension capabilities by integrating two well-established technologies—vision encoder and large language model (LLM). 
The process of using MLLM starts with an input image \(X_v\) and a question prompt \(X_q\) from multi-turn conversation data (\(X_v^1, X_q^1, X_a^1, ..., X_v^n, X_q^n, X_a^n\)) where the turn number is $n$ and $X_a^i$ is the $i$-th turn's answer.
Fig.\ref{fig:framework}(a) illustrates a classic MLLM architecture~\cite{liu2024LLaVA}, which employs a projection layer \(\mathbf{W}_p\) to align the channel dimension of visual tokens extracted via a pre-trained vision encoder \(g_v\) to language embeddings, as follow:
\begin{equation}
    z = (z_q, \mathbf{W}_p(g_v(X_v))
    \label{eq:prelimi}
\end{equation}
,where $\mathbf{z}$ is the input embedding of MLLM, \(z_q\) is the language instruction embedding convert from \(X_q\) by a word to vector model and \(X_v\) is the input image. 

It is easy to notice that the vision encoder and LLM are pre-trained separately, with the projector being the only newly introduced component. 
% Consequently, the disparity in features between the visual encoder and the LLM results in MLLMs becoming overly reliant on language priors, thereby neglecting the visual aspect.
% NoiseBoost introduces noise perturbation to visual tokens, thereby complicating the visual understanding process and compelling the MLLM to allocate more attention to the visual aspect, reducing its reliance on language priors.
Consequently,  MLLMs take shortcuts to be excessively dependent on language priors, neglecting the visual aspect because of the disparity in features.
NoiseBoost introduces noise perturbation to visual tokens, thereby complicating the visual understanding process and compelling the MLLM to allocate more attention to the visual aspect, reducing its reliance on language priors.

\subsection{Supervised Noise Boosting Fine-tuning}
\label{sec:supervised}

Supervised Fine-tuning(SFT) is a widely used training technology, in Fig.\ref{fig:framework}(a). Given an image and a language prompt, the MLLM model directly predicts the linguistic results autoregressively following LLM's convention.  
Without specialized for multi-modality training, MLLM inherited the characteristics of LLM with over-reliance on language priors, as depicted in the token correlations matrix in Fig.\ref{fig:column_analysis}. 
To help MLLM redistribute attention evenly, we propose a noise feature perturbation method $\phi_v$ to disturb pre-trained visual features. 
Based in Eq.~\ref{eq:prelimi}, the noise feature perturbation can be represented as \(z = z_q + \phi_v(W_p(g_v(X_v)))\), where \(z\) is the perturbed visual tokens. So the SFT loss is as follows:
\begin{equation}
    \mathcal{L}_{sft} = - H(y_w | \phi_v(z), X_q),
    \label{eq:sft}
\end{equation}

where $H$ represents the cross-entropy operation used in SFT.The vision feature disturbance makes it hard for MLLM to discern the visual information and pay more attention to image understanding.  We found that adding Gaussian Noise perturbation as \(\phi_v\) to the projected vision tokens can effectively reduce the overreliance on language tokens.  

\subsection{Reinforcement Noise Boosting Learning}
\label{sec:reinforce}
Reinforcement learning has emerged as an essential technology with the rise in popularity of LLMs~\cite{ouyang2022instructgpt}. However, the direct application of reinforcement learning techniques from LLMs to MLLMs without adaptation has proven to be unstable due to data limitations, as also observed in ~\cite{zhao2023hadpo}. 
To address this, we propose the integration of noise feature perturbation for visual tokens to augment visual understanding and SFT into reinforcement learning to enhance training stability as illustrated in Fig.\ref{fig:framework}(b). 
The noise feature perturbation is added directly to visual tokens but in a different training corpus. 
We employ the DPO~\cite{rafailov2024direct}, a classical reinforcement learning algorithm, to illustrate the loss equation:

% \begin{align}
% \mathcal{L_{DPO}} + \mathcal{L_{SFT}} \nonumber &= - \mathbb{E}_{(x, y_w, y_l) \sim \mathcal{D}} \left[ log\sigma \left( \beta log \frac{\pi_\theta(y_w | \phi_v(X_V), X_q)}{\pi_{ref}(y_w | X_V, X_q)} \right. \right.\\
% &\left. \left. - \beta log \frac{\pi_{\theta}(y_l | \phi_v(X_V), X_q)}{\pi_{ref}(y_l | X_v, X_q)} \right) \right] - H(y_w | \phi_v(X_V), X_q) \label{eq:dpo_sft}
% \end{align}
\begin{equation}
\mathcal{L}_{dpo}  = - \mathbb{E}_{(x, y_w, y_l) \sim \mathcal{D}} \left[ log\sigma \left( \beta log \frac{\pi_\theta(y_w | \phi_v(X_V), X_q)}{\pi_{ref}(y_w | X_V, X_q)} \right. \right.\left. \left. - \beta log \frac{\pi_{\theta}(y_l | \phi_v(X_V), X_q)}{\pi_{ref}(y_l | X_v, X_q)} \right) \right].
\label{eq:dpo_sft}
\end{equation}
In this equation, the random feature perturbation function $\phi_v$ is incorporated into the projected visual tokens $\phi_v(X_v)$. The variables $y_w$ and $y_l$ represent the preferred and less preferred outputs, respectively. The model's objective is to maximize the probability of the preferred output and minimize that of the less preferred one. The function $\pi_{\theta}$ denotes the model's policy, while $\pi_{ref}$ signifies the reference policy. The sigmoid function $\sigma$, compresses its input into the range (0, 1), and $\beta$ is a temperature parameter controlling the distribution's sharpness. 
The final reinforcement loss is defined as \( \mathcal{L}_{rl} = \mathcal{L}_{sft} + \mathcal{L}_{dpo}\). In experiments, we observed that a larger noise perturbation on $y_w$ and less on $y_l$ resulted in superior performance. This aligns with our intuition that a challenging visual feature enables MLLM to learn a better attention weight distribution, which the less preferred output $y_l$ does not need.

\subsection{Semi-Supervised Noise Boosting Learning}
\label{sec:semisup}
Semi-supervised learning is a mature technology for mining the ability of unlabeled data but has not been applied to MLLMs. 
The reason is that MLLM's training strategy prevents it from creating a weak augmentation for pseudo labels generation and a strong augmentation for consistency regularization. 
Particularly, MLLMs have no visual augmentation methods, with the assumption that pixel-level image disturbance can mislead understanding of the image content. 
Thanks to NoiseBoost, we can incorporate noise feature perturbations to provide weak and strong distortions for MLLMs without affecting visual understanding and comply with semi-supervised mechanism at the same time. 
The unlabeled loss for semi-supervised learning is 
\begin{equation}
\mathcal{L}_{u}=\frac{1}{\mu B} \sum_{i=1}^{\mu B}\vmathbb{1}(\max(\pi_{ref}(X_v, X_q)) \geq t)H(\hat{q_i}, \pi_{\theta}(\phi_v(X_v), X_q))
\label{eq:pseudolabel}
\end{equation}
,where the reference model $\pi_{ref}$ generate pseudo labels without feature distortion and $\pi_{\theta}$ keep training on noise distorted data for consistency regularization. $t$ is the threshold for filtering noisy pseudo labels, $\mu$ is the ratio of label and unlabeled data, $B$ is the batch size and $\hat{q}_i = arg max(\pi_{ref}(X_v, X_q))$ for artificial labels generation. The final semi-supervised loss is $\mathcal{L}_{semi} = \mathcal{L}_{u} + \mathcal{L}_{sft}$. With NoiseBoost, we can mine the unlabeled data power without waiting for labor-intensive trillions of data cleaning.

% \subsection{Analysis of Perturbation on Gradients}
% [WHICH]    

\section{Experiments}
% \usepackage{graphicx}

\subsection{Setup}

% \usepackage{booktabs}

\begin{table*}[t!]
\small
\centering
\caption{Supervised Fine-tuning results. NoiseBoost consistently improves the performance of MLLM on hallucinated and question-answer datasets. For the caption dataset Flickr30k, we achieve comparable performance since the traditional caption dataset cannot manifest the long, dense captions generated by NoiseBoost. * means trained on our collected dataset. }
\begin{adjustbox}{width=\linewidth}
	\begin{tabular}{l| c| cc c c c c c c}
	\toprule
  \makecell*[c]{Method} &\makecell*[c]{Backbone}
        &\makecell*[c]{POPE} &\makecell*[c]{GQA}  &\makecell*[c]{VixWiz} &\makecell*[c]{Text-\\VQA}
        &\makecell*[c]{MME}
        &\makecell*[c]{SEED\\Bench}&\makecell*[c]{Flickr30K} \\
		\cmidrule(lr){1-1} \cmidrule(lr){2-3} 
  \cmidrule(lr){3-9}
        \color{gray}{\textit{Existing Methods}}&&\\
Flamingo & 80B & -  & -  & 31.6 & -  & -  & -  & 67.2 \\
VLIP-2 & Vicuna-13B & -  & 32.3 & 19.6 & -  & -  & -  & 71.6 \\
InstructBLIP & Vicuna-13B & -  & 49.5 & 33.4 & -  & -  & -  & 82.8 \\
mPLUG-Owl2 & -  & -  & 56.1 & 54.5 & -  & -  & -  & 85.1 \\
Qwen-VL & Qwen-7B & -  & 59.3 & 35.2 & -  & -  & -  & 85.8 \\
Qwen-VL-Chat & Qwen-7B & -  & 57.5 & 38.9 & -  & -  & -  & 81.0 \\
LLaVA-1.5 & Llama-13B & 87.1 & 63.3 & 56.6 & 48.69 & 1523 & 68.2 & 79.5 \\
LLaVA-1.5 & Llama-7B & 86.9 & 61.9 & 54.3 & 46.07 & 1507 & 66.2 & 74.9 \\
    \cmidrule(lr){1-1}\cmidrule(lr){2-2}\cmidrule(lr){3-9}
    \color{gray}{\textit{NoiseBoost}}&&\\
% QwenVL* & Qwen7B & 0.09 & 40.52 & 4.32 & 61.72 & 1547 & 33.5 & 12.3 \\ 
% + NoiseBoost & Qwen7B & -  & -  & -  & -  & -  & -  &  \\
% \cmidrule(lr){1-1}\cmidrule(lr){2-2}\cmidrule(lr){3-9}
LLaVA-1.5* & Llama-13B & 88.3 & 64.0 & 59.8 & 49.5 & 1540 & \textbf{69.2} & \textbf{81.2} \\
+ NoiseBoost & Llama-13B & \textbf{88.4}  & \textbf{64.2}  & \textbf{61.5}  & \textbf{49.8}  & \textbf{1580}  & 69.1  & 80.8 \\ 
    \cmidrule(lr){1-1}\cmidrule(lr){2-2}\cmidrule(lr){3-9}

    LLaVA-1.5* & Llama-7B & 87.2 & 62.3 & 54.6 & 47.18 & 1501 & 66.9 & 73.3 \\
+ NoiseBoost & Llama-7B & \textbf{88.4} & \textbf{63.4} & \textbf{57.1} & \textbf{47.8} & \textbf{1531} & \textbf{67.7} & \textbf{73.8} \\
	\bottomrule
	\end{tabular}
\end{adjustbox}
 \label{tab:main_result}
\end{table*}

\textbf{Baseline and Data.} 
We have chosen LLaVA-1.5~\cite{liu2024LLaVA} and QwenVL~\cite{bai2023qwenvl}, two recently released state-of-the-art MLLMs, as our baseline. In addition to the data incorporated in LLaVA-1.5, we have supplemented our dataset with COCO captions and ShareGPT4v~\cite{chen2023sharegpt4v} to reach a total of 800K entries, thereby matching the performance of LLaVA-1.5 because thousands of images could not be downloaded from the original 665K meticulously curated dataset~\cite{liu2023LLaVA1.5}. The data serves as a comparable baseline for LLaVA-1.5-7B, but it falls short for QwenVL-Chat~\cite{bai2023qwenvl}. 
Given that QwenVL~\cite{bai2023qwenvl} has not released its data for retraining, we only evaluate QwenVL for human evaluation to establish a relatively fair comparison. 
Reinforcement learning datasets for MLLM are limited, we use HA-DPO dataset ~\cite{zhao2023hadpo} which contains 18k images. Although the dataset is limited in scale, NoiseBoost can also achieve performance gain compared to previous methods. Semi-supervised learning datasets are constructed by splitting the SFT data into 30\% and 50\% with others used for unlabeled data learning.
In summary, with the data managed to maintain a fair comparison, we tested NoiseBoost on LLaVA-1.5 using Llama7B and Llama13B to assess backbone generalizability and on QwenVL to evaluate performance across different LLM styles. 
% Because training QwenVL with our data degrades its performance, we only evaluate caption capability by human evaluations

\textbf{Implementation Details.}
For supervised fine-tuning, We use a batch size of 192 with accumulation steps setting to 2 for training, similar to ~\cite{liu2023LLaVA1.5}, with 24 V100 training for 16 hours, about 384 GPU hours. For reinforcement learning and semi-supervised learning 30\% setting, the training only needs around 90 GPU hours and 150 GPU hours to finish because of not much data. We set the learning rate to 2e-5 in for SFT and semi-supervised learning. Reinforcement learning uses a small learning rate 2e-6 because of data deficiency resulting in unstable training. The weight decay and warmup ratio are set to 0.0 and 0.03 respectively. The model length is 2048, the same as ~\cite{liu2023LLaVA1.5} but 1024 for QwenVL~\cite{bai2023qwenvl} for fewer training hours. All of our experiments are conducted on float16 with deepspeed due to GPU memory limitations. For noise perturbation, we set the noise scale to 0.5 with a 50\% chance of triggering perturbation for all experiments if not specified without tuning the parameters. 

\textbf{Evaluations.}
We conduct an evaluation of various datasets, including the hallucination dataset POPE~\cite{li2023pope}, question-answer datasets~\cite{hudson2019gqa,gurari2018vizwiz,li2023mme,li2023seed}, and the caption dataset~\cite{plummer2015flickr30k}. Given the inherent difficulty in assessing long captions using automated tools, we supplement our study with a collection of 1,000 images for human evaluation. For automated evaluations, we utilized ~\cite{lmms_eval2024}, a publicly available tool designed to facilitate the evaluation of all MLLM datasets. However, we observed suboptimal performance when assessing QwenVL-Chat using ~\cite{lmms_eval2024}, attributable to a minor modification in the evaluation prompt can lead to significant discrepancies in the MLLM results. To ensure a fair evaluation, we maintained uniformity in all evaluations, refraining from prompt tuning for a single model and only evaluating QwenVL on human evaluation. For human-labeled captions, we ask annotators to select reasons for captioning results other than the binary right or wrong.
% For a fair evaluation,  we also evaluate with the tools provided by ~\cite{li2023monkey}. When evaluated by humans, we ask annotators to select the detailed reasons if the caption response is wrong. 

\subsection{Quantative Experiments}
In this section, we analyze NoiseBoost's performance gain in SFT, reinforcement learning and semi-supervised learning.

\begin{figure}
    \centering
    \includegraphics[width=1.0\textwidth]{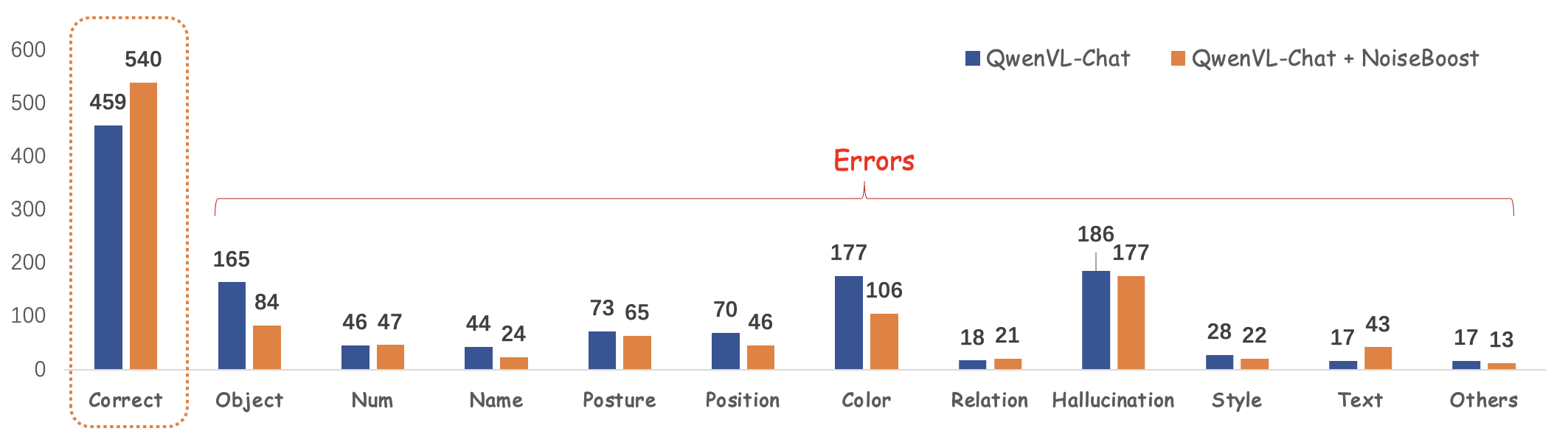}
    \caption{Human evaluation of dense captions for 1k images with prompt "Describe the image and its style in detail". The correct category means the totally accurate image with other categories are errors identified by human annotators. NoiseBoost consistently reduce error on nearly all categories.}
    \label{fig:human_eval}
\end{figure}

\textbf{Supervised Fine-tuning.}
We conduct experiments on LLaVA-1.5 7B/13B and QwenVL to test variations in backbone and architecture. As demonstrated in Tab.\ref{tab:main_result}, NoiseBoost consistently enhances performance across nearly all datasets, with gains exceeding 1\% over most datasets, no matter whether the datasets are hallucination-based or question-answer-based. 
For the LLaVA-1.5 with the Llama 13B model, the MME~\cite{li2023mme} reached 1580, 40 points higher than the original model. 
Notably, we only achieve performance comparable to the baseline on Flickr30K~\cite{plummer2015flickr30k} because NoistBoost is more likely to generate rich caption data, which differs from traditional caption evaluation datasets. 
However, most MLLM automatic evaluations are notorious for not aligning with human feelings.

We further assessed the dense caption performance using human labeling, as depicted in Fig. \ref{fig:human_eval}. NoiseBoost achieves an accuracy of 540/1000, which is 8.1\% higher than the QwenVL-Chat baseline. With human labeling, error categories are classified. A detailed explanation of each category can be found in the supplementary material Sup.~\ref{sup:eval_guid}. Our improvements primarily stem from object error and hallucinations, which refer to the description of erroneous objects or non-existent objects, respectively. The results indicate that noise feature perturbation can redistribute the attention weights of the MLLM, leading to more pronounced improvements in object-related information in the image.

\begin{table}
\centering
\caption{Reinforcement learning result. NoiseBoost consistently improves over all datasets.}
\begin{adjustbox}{width=0.75\linewidth}
\begin{tabular}{@{}l|cccccc@{}}
\toprule
Model & POPE & GQA & VizWiz & MME & SeedBench & ScienceQA \\ \midrule
LLaVA-1.5 DPO & 86.3 & 60.1 & 53.9 & 1516 & 66.3 & 66.9 \\ 
\rowcolor[HTML]{EFEFEF} + NoiseBoost & \textbf{87.2} & \textbf{61.8} & \textbf{54.7} & \textbf{1528} & \textbf{66.5} & \textbf{70.3} \\ \bottomrule
\end{tabular}
\end{adjustbox}
\label{tab:dpo_train}
\end{table}

\begin{wrapfigure}{r}{0.5\textwidth}
  \vspace{-10pt}
  \centering
  \includegraphics[width=0.5\textwidth]{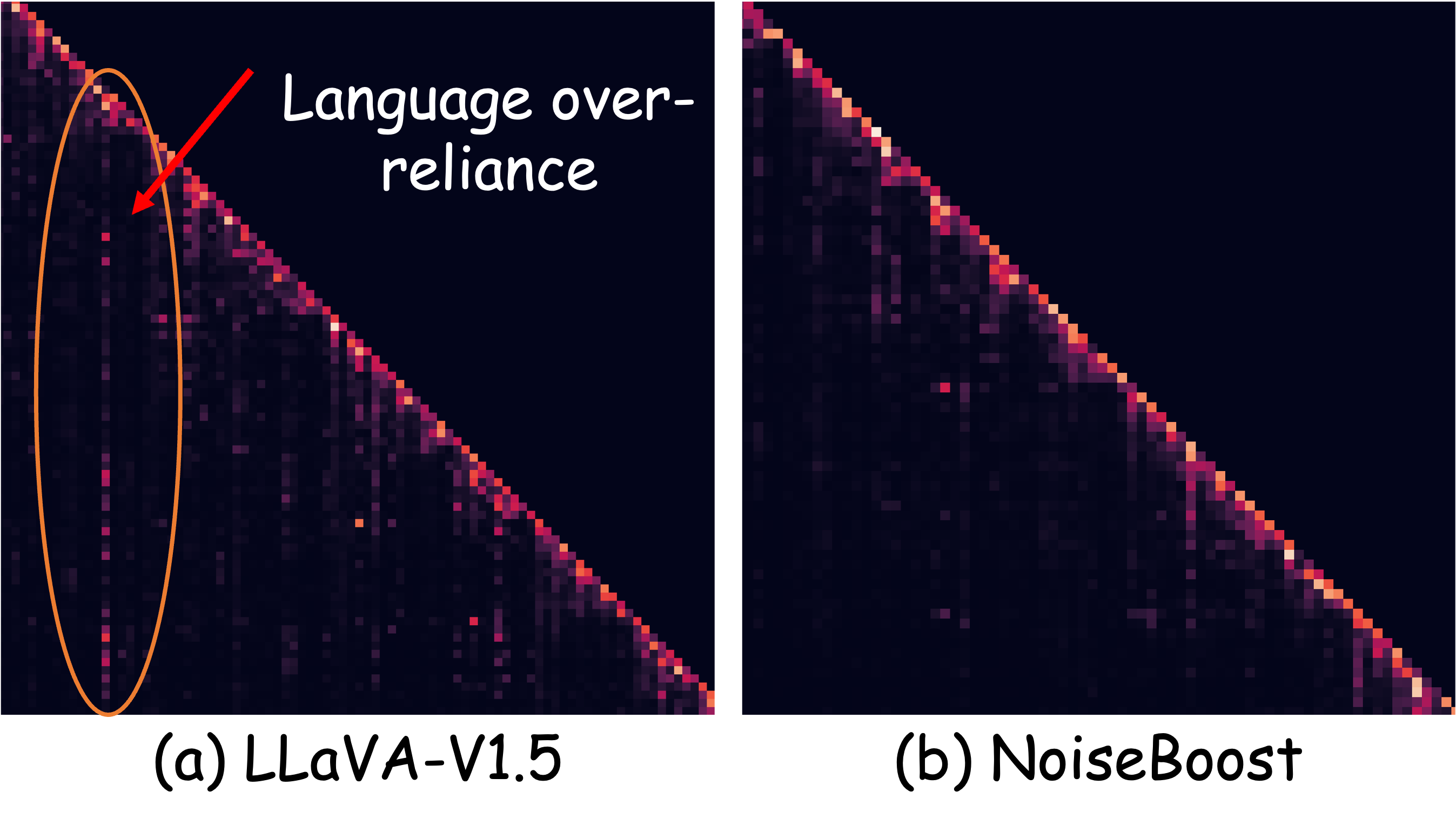}
  \caption{Analysis of the cause of hallucination is the over-reliance on language tokens circled in read which NoiseBoost doesn't have.}
  \label{fig:column_analysis}
  \vspace{-10pt}
\end{wrapfigure}
\textbf{Reinforcement Learning.}
To align the behaviour of MLLM with actual human responses, DPO~\cite{rafailov2024direct} serves as a prevalent reinforcement learning technique that requires only paired data for training. However, the DPO is first proposed in LLM and has no adaptation for MLLMs. We inject the noise perturbation to the preferred visual features with the assumption that harder consistency learning achieves better results.
Upon testing with HA-DPO~\cite{zhao2023hadpo} as shown in Table~\ref{tab:dpo_train}, NoiseBoost improve ScienceQA with 3.4\% and consistently enhances performance by approximately 1\% on both the hallucination dataset~\cite{li2023pope} and various question-answer datasets~\cite{hudson2019gqa,gurari2018vizwiz,li2023mme,li2023seed,saikh2022scienceqa}. However, it is noteworthy that the degree of improvement is relatively less in comparison to SFT. This can be attributed to two primary factors: the limited scale of HA-DPO, which restricts the full potential of NoiseBoost due to fewer training steps, and the lack of proper tuning of the noise scale injected into the features, which prevents a fair comparison.

% \usepackage{graphicx}
\begin{table}
\centering
\caption{Semi-supervised learning experiments, NoiseBooost enables MLLM mining unlabeled data and achieve similar performance with 50\% data.}
\label{tab:semi_suptrain}
\begin{adjustbox}{width=0.85\linewidth}
\begin{tabular}{lcccccc}
\hline
 & POPE & GQA & VizWiz & MME & Seedbench & ScienceQA \\ \hline
30\% Data & 86.0 & 60.3 & 44.1 & 1426 & 67.0 & 67.9 \\
\rowcolor[HTML]{EFEFEF} 
+ NoiseBoost & \textbf{87.4} & \textbf{62.5} & \textbf{54.9} & \textbf{1509} & \textbf{67.2} & \textbf{69.1} \\
50\% Data & 86.9 & 62.4 & 54.3 & 1490 & \textbf{66.8} & 70.0 \\
\rowcolor[HTML]{EFEFEF} 
+ NoiseBoost & \textbf{88.0} & \textbf{62.5} & \textbf{55.2} & \textbf{1553} & 67.0 & \textbf{71.0} \\
\bottomrule
\end{tabular}%
\end{adjustbox}
\end{table}
\textbf{Semi-Supervised Learning.}
To unleash the power of unlabeled data, we incorporate NoiseBoost to create teacher-student architecture as in MeanTeacher~\cite{tarvainen2017mean}, which is a classic semi-supervised learning technique. The teacher produces pseudo labels, and the student learns with strong augmented images for consistency regularization. With noise perturbation, we inject Gaussian noise during student learning and keep the original model frozen as a teacher. The experiments in Tab.\ref{tab:semi_suptrain} show that LLaVA-1.5 can reach similar performance with only 50\% of the data, which sheds light on mining the power of unlabeled data.

\subsection{Qalitative Experiments}
\begin{figure}
    \centering
    \includegraphics[width=\linewidth]{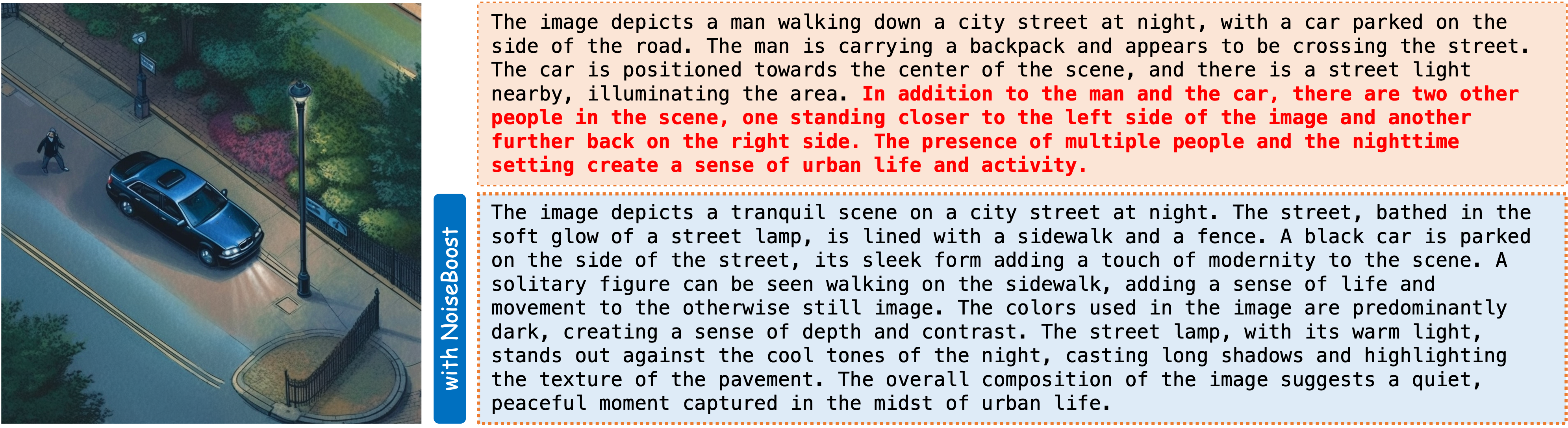}
    \caption{Qualitative evaluation shows that NoiseBoost can generate long detailed captions without hallucinations.}
    \label{fig:qualitative}
\end{figure}

We conduct a series of experiments using a street image, which is prone to cause hallucinations of "people" due to the common association of people walking in streets in language. As illustrated in Fig.~\ref{fig:qualitative}, the original model tends to hallucinate during response generation, primarily due to an over-reliance on language priors. To substantiate our hypothesis, we visualized the token correlation map, using the final layer attention maps from LLM's last token generation. The column attention, highlighted in red in Fig.~\ref{fig:column_analysis} (a), indicates that the subsequently generated tokens are overly dependent on a specific language anchor token, leading to a neglect of visual tokens. 
The column phenomenon emerges in the middle, coinciding with the occurrence of hallucination. 
The tendency for hallucination becomes severe during the generation of longer responses. 
NoiseBoost, however, disrupts the overconfidence in specific hallucination tokens and is capable of generating extended captions without errors.
After the introduction of noise feature perturbation, the LLM redistributes attention weights more evenly, as shown in Fig.~\ref{fig:column_analysis} (b).

\section{Ablations}
\textbf{Different Feature Perturbation Scale.}
We choose Gaussian noise with upper and lower bound $[0, 1]$ in our paper. To test the robustness of feature perturbation on different scales and probabilities, we conduct extensive experiments. As in Tab.\ref{tab:noise_prob}, NoiseBoost is robust to the scale of noise-injected with not much variation with changing the hyperparameters, duo to page limits see full table in Sup.\ref{sup:tab_noise_prob}. An interesting phenomenon is that with the increasing of noise scale, the performance first 
increase and then decrease, which can be explained by the fact that the MLLM training process needs noise to break the language reliance, but too much noise can harm the learning process.

% \usepackage{graphicx}
\begin{table}
\centering
\small
\caption{Different noise probability and noise scale. With an increase in noise prob and scale, the MLLM's performance is robust, but too much noise may affect the learning process. }
\label{tab:noise_prob}
\begin{adjustbox}{width=0.85\linewidth}
\begin{tabular}{cc|cccccccc}
\toprule
\begin{tabular}[c]{@{}c@{}}Noise\\ Prob\end{tabular} & \begin{tabular}[c]{@{}c@{}}Noise\\ Scale\end{tabular} & POPE & GQA & \begin{tabular}[c]{@{}c@{}}Viz\\ Wiz\end{tabular} & \begin{tabular}[c]{@{}c@{}}Text\\ VQA\end{tabular} & \begin{tabular}[c]{@{}c@{}}Seed\\ bench\end{tabular} & MME & \begin{tabular}[c]{@{}c@{}}Text\\ Caps\end{tabular} & \begin{tabular}[c]{@{}c@{}}Flickr\\ 30K\end{tabular} \\ \hline
0 & 0 & 87.2 & 62.3 & 54.6 & 47.6 & 66.9 & 1501 & 96.9 & 73.3 \\
% \rowcolor[HTML]{EFEFEF} 
0.1 & 0.1 & 88.1 & 63.4 & 56.4 & \textbf{47.9} & 67.2 & 1506 & 98.4 & 73.1 \\
% 0.3 & 0.5 & 88.0 & 63.1 & 54.5 & 47.4 & 66.8 & 1517 & 98.9 & 72.8 \\
% \rowcolor[HTML]{EFEFEF} 
0.5 & 0.1 & 88.2 & \textbf{63.4} & 54.4 & 47.5 & 66.9 & 1504 & 97.2 & 73.2 \\
% 0.5 & 0.3 & 88.2 & 63.2 & 54.0 & 47.5 & 67.0 & 1522 & 97.9 & 72.8 \\
% \rowcolor[HTML]{EFEFEF} 
0.5 & 0.5 & \textbf{88.4} & \textbf{63.4} & \textbf{57.1} & 47.8 & \textbf{67.7} & 1531 & \textbf{100.6} & \textbf{73.8} \\
% 0.5 & 0.7 & 88.1 & 63.2 & 55.4 & 47.2 & 66.9 & 1525 & 98.4 & 73.2 \\
% \rowcolor[HTML]{EFEFEF} 
0.5 & 0.9 & 87.9 & 63.0 & 55.2 & 47.0 & 66.6 & 1517 & 98.6 & 73.1 \\
0.7 & 0.5 & 87.8 & 63.0 & 54.0 & 47.7 & 66.6 & \textbf{1532} & 96.8 & 72.3 \\
% \rowcolor[HTML]{EFEFEF} 
% 0.9 & 0.1 & 88.3 & 63.2 & \textbf{57.1} & 47.7 & 67.1 & 1524 & 97.8 & 72.2 \\
0.9 & 0.5 & 87.9 & 62.9 & 55.8 & 47.1 & 66.8 & 1522 & 96.8 & 72.7 \\ 
\bottomrule
\end{tabular}%
\end{adjustbox}
\end{table}

\textbf{Other Perturbation Methods.}
% \usepackage{booktabs}
\begin{table}
\centering
\caption{Comparision with pixel level and feature level perturbations.}
\begin{adjustbox}{width=0.8\linewidth}
\begin{tabular}{@{}l|ccccccc@{}}
\toprule
\begin{tabular}[c]{@{}l@{}}Perturbation\\ Methods\end{tabular} & POPE & GQA & \begin{tabular}[c]{@{}c@{}}Scien\\ ceQA\end{tabular} & \begin{tabular}[c]{@{}c@{}}Viz\\ Wiz\end{tabular} & MME & \begin{tabular}[c]{@{}c@{}}Seed\\ bench\end{tabular} & \begin{tabular}[c]{@{}c@{}}Flickr\\ 30K\end{tabular} \\ \midrule
ColorJitter & \textbf{88.6} & 62.5 & \textbf{69.9} & \textbf{58.7} & 1485 & 67.3 & 70.5 \\ 
% \rowcolor[HTML]{EFEFEF} 
RandomCrop & 86.4 & 63.1 & 69.5 & 56.6 & 1516 & 66.7 & \textbf{74.6} \\ 
GaussianNoise & 86.0 & 62.4 & 69.5 & 55.4 & 1507 & 65.4 & 70.7 \\ \midrule
Dropout & 88.0 & 63.3 & 69.7 & 55.8 & 1489 & 67.3 & 71.4 \\ 
% \rowcolor[HTML]{EFEFEF} 
Dropout + NoiseBoost & 88.1 & 63.3 & 69.5 & 57.9 & 1491 & 67.4 & 74.3 \\ 
% \hline
\rowcolor[HTML]{EFEFEF} 
Ours & 88.4 & \textbf{63.4} & \textbf{69.9} & 57.1 & \textbf{1531} & \textbf{67.7} & 73.8 \\ \bottomrule
\end{tabular}
\label{tab:other_augs}
\end{adjustbox}
\end{table}
Perturbation methods can be broadly classified into pixel-level and feature-level categories. In the case of pixel-level perturbations, we evaluate the efficacy of conventional image augmentations. For feature-level perturbations, we opt for dropout as a comparative measure, given the absence of alternative feature augmentation techniques.

As demonstrated in Table~\ref{tab:other_augs}, pixel-level distortions such as RandomCrop and GaussianNoise induce more hallucinations in POPE~\cite{li2023pope}, as these distortions impact the appearance or even the existence of the object. ColorJitter, which solely alters the image's colour, does not increase incorrect object hallucinations in POPE, but it does degrade performance in visual understanding datasets like MME~\cite{li2023mme} due to the disparity in colour. Pixel-level distortions, therefore, either crop images, induce object hallucinations, or disrupt the colour space, thereby affecting visual comprehension.

Since the inception of the Deep Learning era, Dropout has been a widely used feature perturbation method. We incorporate dropout into visual features post-projection, akin to NoiseBoost, and adhere to the convention by setting the dropout rate at 0.1. As per Table~\ref{tab:other_augs}, Dropout only yields performance comparable to the baseline, which can be attributed to the fact that MLLM training methods already employ this technology for backbone training. The performance can be enhanced with NoiseBoost, thereby validating the effectiveness of our method.

\textbf{Feature Perturbation on Language.}
% \usepackage{graphicx}
\begin{table}
\centering
\caption{Comparision with noise perturbation on language tokens. }
% \resizebox{\columnwidth}{!}{%
\begin{tabular}{l|ccccccc}
\hline
 & POPE & GQA & ScienceQA & VizWiz & TextVQA & MME & Seedbench \\ \hline
+ Lan & 87.9 & 61.1 & 59.7 & 46.6 & 45.1 & 1477 & 65.1 \\
% \rowcolor[HTML]{EFEFEF}
+ Lan Vis & 88.1 & 62.7 & 66.8 & 56.3 & 46.4 & 1507 & 67.1 \\ 
Ours & \textbf{88.4} & \textbf{63.4} & \textbf{69.9} & \textbf{57.1} & \textbf{47.8} & \textbf{1531} & \textbf{67.7}  \\ \hline
\end{tabular}%
% }
\label{tab:other_noise}
\end{table}
NoistBoost only adds feature perturbation to visual features to align with LLM feature space and break the over-reliance on language priors. We also study whether the perturbation is effective vice versa. By adding noise to language embeddings before LLM backbone, we found a performance degradation among nearly all benchmarks. From Tab.\ref{tab:other_noise}, we can conclude that the foundation LLM has a strong pre-trained knowledge, which should not be affected during training. However, with visual token noise perturbation in NioseBoost, the performance can also be enhanced.

\subsection{Limitations and Societal Impacts}
NoiseBoost serves as a fundamental method capable of mitigating the hallucination phenomenon in MLLM throughout all stages of training. The feature perturbation technique employed by NoiseBoost is a rudimentary training strategy that not only avoids negative societal impacts but also propels the advancement of multi-modal AI assistants. However, it is important to note that while NoiseBoost does not necessitate any additional costs or modifications to the MLLM structure, it doesn't change the existing methods. Presently, MLLM incorporates large language models without any module resembling the human brain, which should be developed at the architecture level.

\section{Conclusions}
Recent advancements in MLLM have been swift, yet these models can induce hallucinations, thereby limiting their practical applications. This paper introduces a simple, broadly applicable method, termed NoiseBoost, designed to enhance visual comprehension and mitigate hallucinations in MLLM without incurring additional costs. Specifically, NoiseBoost incorporates Gaussian noise into visual tokens to diminish the excessive reliance on language priors, a characteristic inherited from LLMs. Through comprehensive experimentation, we demonstrate that feature perturbation can augment MLLM performance without extra expenditure, and that NoiseBoost currently stands as the most efficacious feature perturbation technique. Moreover, we equip MLLM with semi-supervised learning capabilities by employing NoiseBoost to establish teacher-student networks. Collectively, we posit that NoiseBoost can serve as a fundamental method for training MLLMs and illuminate the path towards exploiting unlabeled data for large language models.

\bibliography{cite_2024}
\bibliographystyle{unsrt}
% \bibliographystyle{plain} 
% {
% 	\small
% 	\bibliography{cite_2024}
% 	% \nocite{*}
% }

%%%%%%%%%%%%%%%%%%%%%%%%%%%%%%%%%%%%%%%%%%%%%%%%%%%%%%%%%%%%
\renewcommand\thefigure{A\arabic{figure}}
\renewcommand\thetable{A\arabic{table}}  
\renewcommand\theequation{A\arabic{equation}}
\setcounter{equation}{0}
\setcounter{table}{0}
\setcounter{figure}{0}

\newpage

\appendix
\section{Appendix}

\subsection{Human Evaluation Guidance for Dense Captions.}
\label{sup:eval_guid}
To align the MLLM's evaluation with human preferences, we ask human annotators to evaluate the dense captions with detailed error category labeling, including errors about object, number, name, posture, position, color, relation, hallucination, style, text, and others.

\textbf{Object Error.} Errors of object descriptions. For example, describing a phone when it's actually an iPad, or describing short hair as long hair, short sleeves as padded jackets

\textbf{Number Error.} Errors of number.
For example, if there are two people dancing in the picture, the MLLM says three people.

\textbf{Name Error.} Error for proper noun. Such as incorrect descriptions of a person's name, place of interest, or idiom.

\textbf{Posture Error.} Errors of posture or movement. The object is not doing the described action.

\textbf{Position Error.} Errors of object position. The object is not in the described position, such as in the picture's top, bottom, left, or right.

\textbf{Color Error.} Errors of color descriptions.

\textbf{Relation Error.} Errors for relations among subjects. such as the description of "two people, one on the other's shoulder," but in the image is a left-right or front-back relationship.

\textbf{Hallucination Error.} Error of hallucinations. The object described in the picture does not exist.

\textbf{Style Error.} The described style is wrong, such as light black and white, the actual picture is colorful and heavy, etc.

\textbf{Text Error.} Error of the text descriptions in the image. 

\textbf{Other Errors.} Errors not listed in the above such as repetition of MLLM response. 

\textbf{Correct.} The descriptions not labeled into any categories of errors are considered correct. 

\subsection{Noise Perturbation Scale and Probability Variations}
The full table shows changing the scale and probability of noise feature perturbation for NioseBoost. We found that NoiseBoost is robust with the variation of hyperparameters. 
\begin{table}[ht]
\centering
\caption{Different noise probability and noise scale. With an increase in noise prob and scale, the MLLM's performance is robust, but too much noise may affect the learning process. }
\label{sup:tab_noise_prob}
\begin{adjustbox}{width=0.85\linewidth}
\begin{tabular}{cc|cccccccc}
\toprule
\begin{tabular}[c]{@{}c@{}}Noise\\ Prob\end{tabular} & \begin{tabular}[c]{@{}c@{}}Noise\\ Scale\end{tabular} & POPE & GQA & \begin{tabular}[c]{@{}c@{}}Viz\\ Wiz\end{tabular} & \begin{tabular}[c]{@{}c@{}}Text\\ VQA\end{tabular} & \begin{tabular}[c]{@{}c@{}}Seed\\ bench\end{tabular} & MME & \begin{tabular}[c]{@{}c@{}}Text\\ Caps\end{tabular} & \begin{tabular}[c]{@{}c@{}}Flickr\\ 30K\end{tabular} \\ \hline
0 & 0 & 87.2 & 62.3 & 54.6 & 47.6 & 66.9 & 1501 & 96.9 & 73.3 \\
% \rowcolor[HTML]{EFEFEF} 
0.1 & 0.1 & 88.1 & 63.4 & 56.4 & \textbf{47.9} & 67.2 & 1506 & 98.4 & 73.1 \\
0.3 & 0.5 & 88.0 & 63.1 & 54.5 & 47.4 & 66.8 & 1517 & 98.9 & 72.8 \\
% \rowcolor[HTML]{EFEFEF} 
0.5 & 0.1 & 88.2 & \textbf{63.4} & 54.4 & 47.5 & 66.9 & 1504 & 97.2 & 73.2 \\
0.5 & 0.3 & 88.2 & 63.2 & 54.0 & 47.5 & 67.0 & 1522 & 97.9 & 72.8 \\
% \rowcolor[HTML]{EFEFEF} 
0.5 & 0.5 & \textbf{88.4} & \textbf{63.4} & \textbf{57.1} & 47.8 & \textbf{67.7} & 1531 & \textbf{100.6} & \textbf{73.8} \\
0.5 & 0.7 & 88.1 & 63.2 & 55.4 & 47.2 & 66.9 & 1525 & 98.4 & 73.2 \\
% \rowcolor[HTML]{EFEFEF} 
0.5 & 0.9 & 87.9 & 63.0 & 55.2 & 47.0 & 66.6 & 1517 & 98.6 & 73.1 \\
0.7 & 0.5 & 87.8 & 63.0 & 54.0 & 47.7 & 66.6 & \textbf{1532} & 96.8 & 72.3 \\
% \rowcolor[HTML]{EFEFEF} 
0.9 & 0.1 & 88.3 & 63.2 & \textbf{57.1} & 47.7 & 67.1 & 1524 & 97.8 & 72.2 \\
0.9 & 0.5 & 87.9 & 62.9 & 55.8 & 47.1 & 66.8 & 1522 & 96.8 & 72.7 \\ 
\bottomrule
\end{tabular}%
\end{adjustbox}
\end{table}

\subsection{Hard Cases Qualitative Examples}
We test NoiseBoost on hard captioning cases with various categories. NoiseBoost shows honest performance on test casts. 
\begin{figure}[hbt!]
    \centering
    \includegraphics[width=\linewidth]{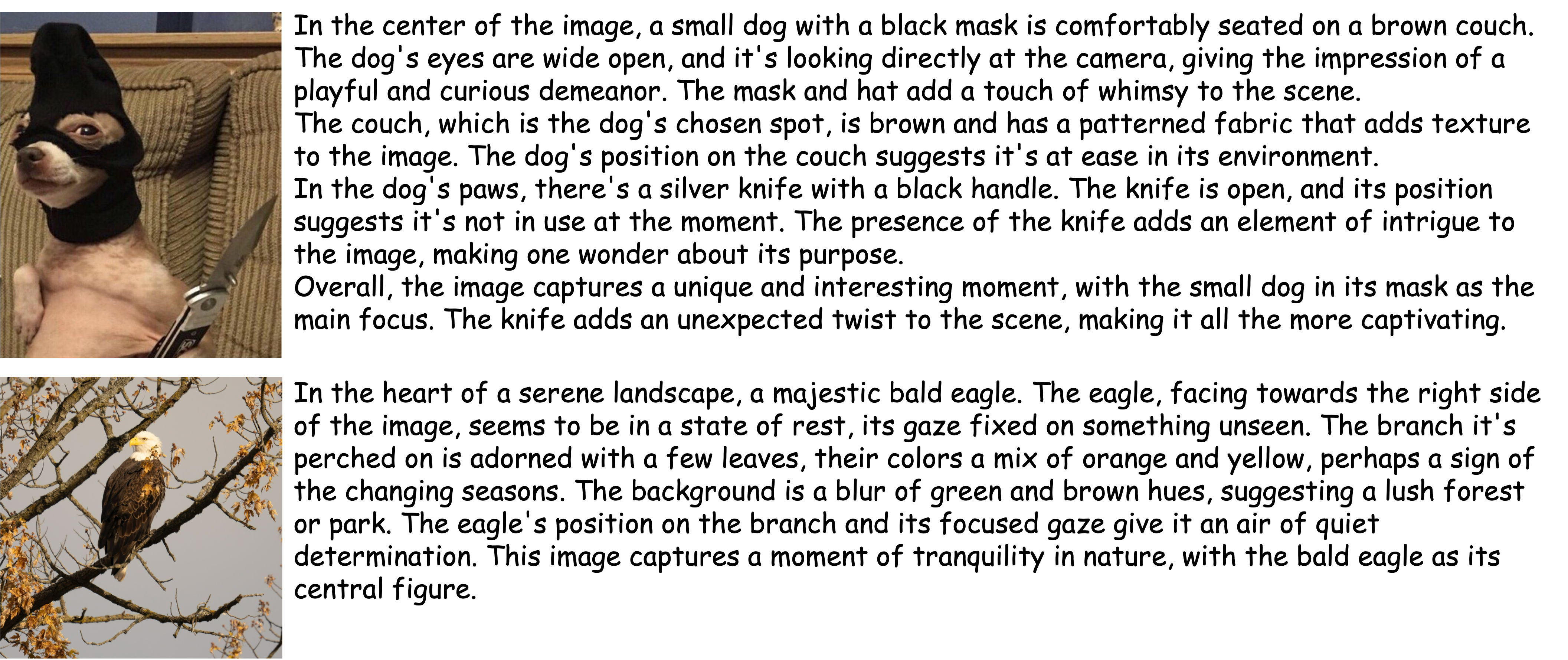}
    \caption{Animal Captioning}
\end{figure}

\begin{figure}[hbt!]
    \centering
    \includegraphics[width=\linewidth]{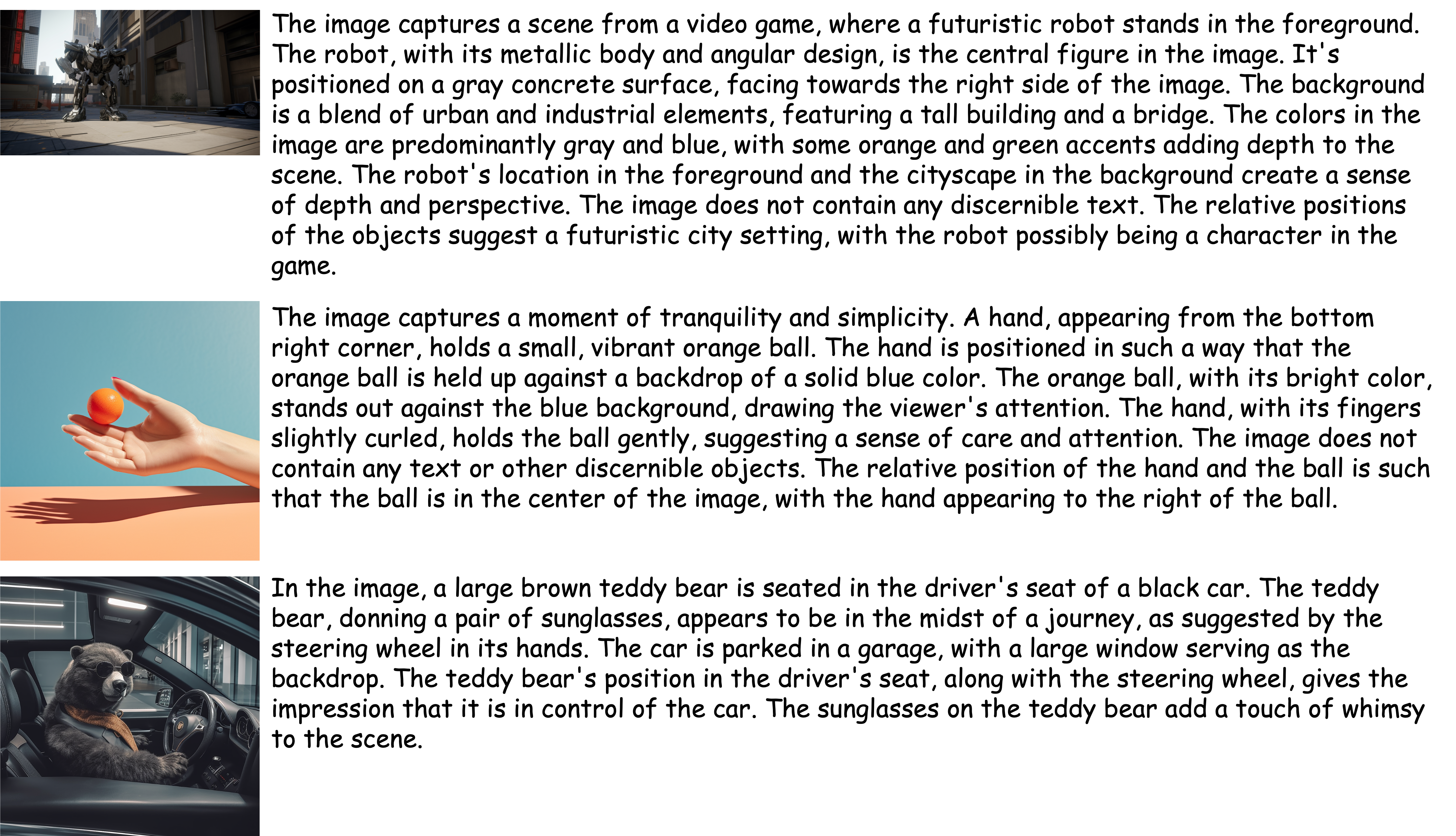}
    \caption{Generated Image Captioning}
\end{figure}

\begin{figure}[hbt!]
    \centering
    \includegraphics[width=\linewidth]{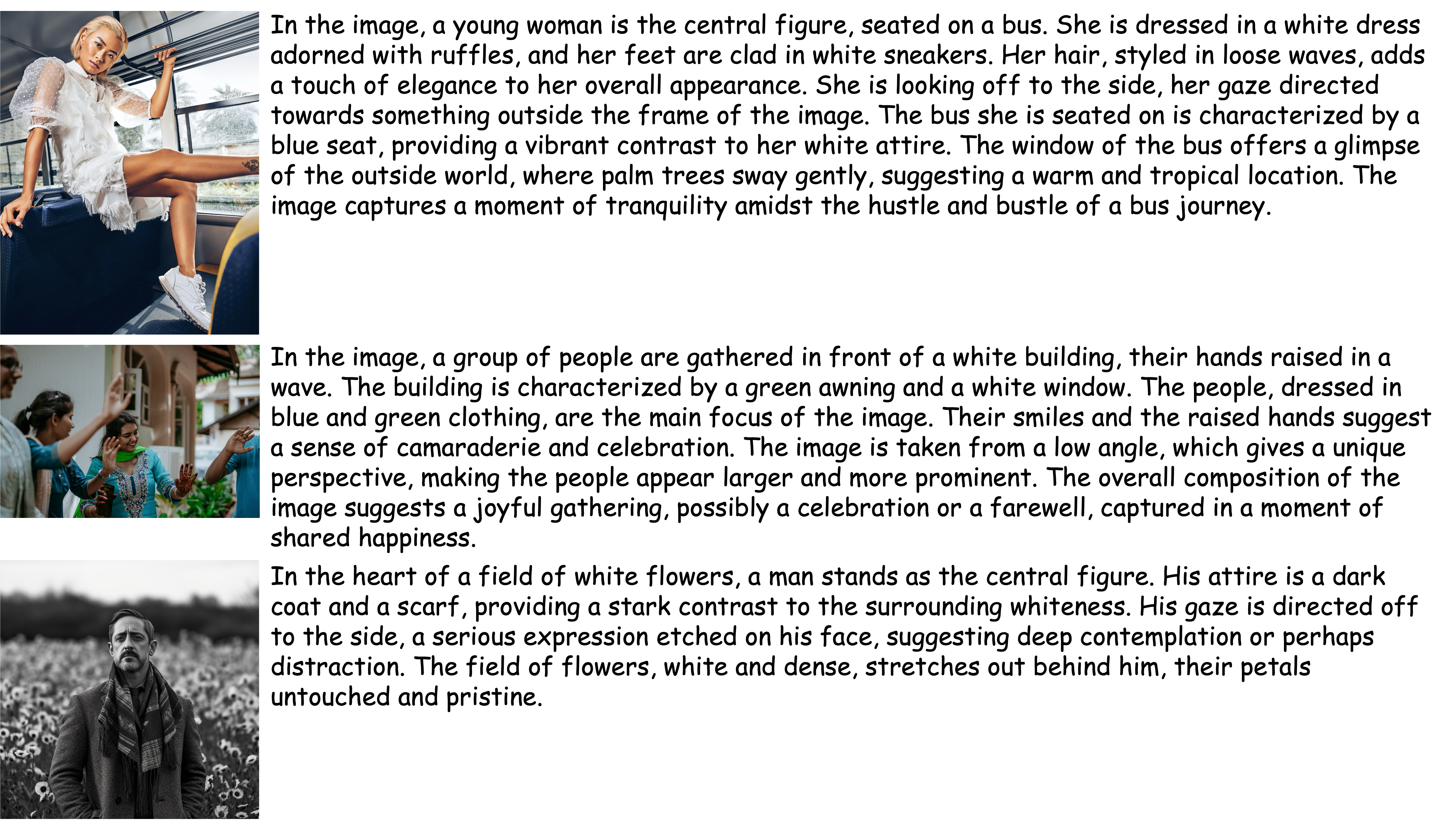}
    \caption{Human Captioning}
\end{figure}

\begin{figure}[hbt!]
    \centering
    \includegraphics[width=\linewidth]{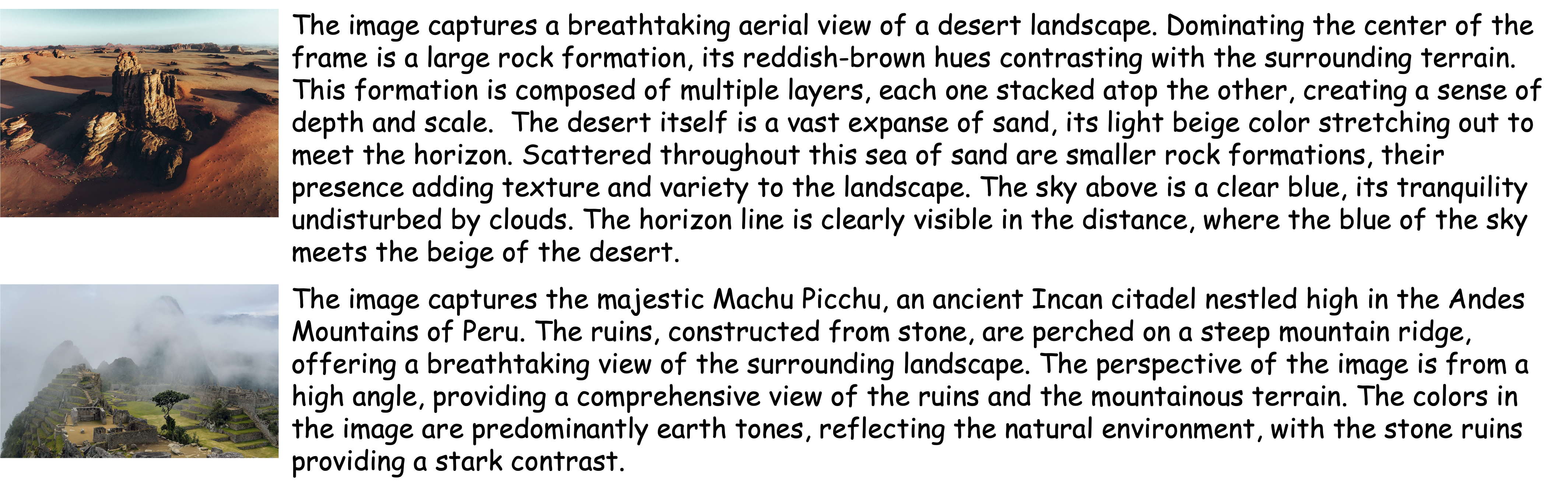}
    \caption{Scenary Captioning}
\end{figure}

\end{document}